# PEGASUS: A policy search method for large MDPs and POMDPs

**Andrew Y. Ng**
Computer Science Division
UC Berkeley
Berkeley, CA 94720

**Michael Jordan**
Computer Science Division
& Department of Statistics
UC Berkeley
Berkeley, CA 94720

## Abstract

We propose a new approach to the problem of searching a space of policies for a Markov decision process (MDP) or a partially observable Markov decision process (POMDP), given a model. Our approach is based on the following observation: Any (PO)MDP can be transformed into an "equivalent" POMDP in which all state transitions (given the current state and action) are deterministic. This reduces the general problem of policy search to one in which we need only consider POMDPs with deterministic transitions. We give a natural way of estimating the value of all policies in these transformed POMDPs. Policy search is then simply performed by searching for a policy with high estimated value. We also establish conditions under which our value estimates will be good, recovering theoretical results similar to those of Kearns, Mansour and Ng [7], but with "sample complexity" bounds that have only a polynomial rather than exponential dependence on the horizon time. Our method applies to arbitrary POMDPs, including ones with infinite state and action spaces. We also present empirical results for our approach on a small discrete problem, and on a complex continuous state/continuous action problem involving learning to ride a bicycle.

## 1 Introduction

In recent years, there has been growing interest in algorithms for approximate planning in (exponentially or even infinitely) large Markov decision processes (MDPs) and partially observable MDPs (POMDPs). For such large domains, the value and $Q$-functions are sometimes complicated and difficult to approximate, even though there may be simple, compactly representable policies that perform very well. This observation has led to particular interest in *direct policy search* methods (e.g., [16, 8, 15, 1, 7]), which attempt to choose a good policy from some restricted class of policies.

Most approaches to policy search assume access to the POMDP either in the form of the ability to execute trajectories in the POMDP, or in the form of a black-box "generative model" that enables the learner to try actions from arbitrary states. In this paper, we will assume a stronger model than these: roughly, we assume we have an implementation of a generative model, with the difference that it has no internal random number generator, so that it has to ask us to provide it with random numbers whenever it needs them (such as if it needs a source of randomness to draw samples from the POMDP's transition distributions). This small change to a generative model results in what we will call a deterministic simulative model, and makes it surprisingly powerful.

We show how, given a deterministic simulative model, we can reduce the problem of policy search in an arbitrary POMDP to one in which all the transitions are deterministic—that is, a POMDP in which taking an action $a$ in a state $s$ will always deterministically result in transitioning to some fixed state $s'$. (The initial state in this POMDP may still be random.) This reduction is achieved by transforming the original POMDP into an "equivalent" one that has only deterministic transitions.

Our policy search algorithm then operates on these "simplified" transformed POMDPs. We call our method PEGASUS (for Policy Evaluation-of-Goodness And Search Using Scenarios, for reasons that will become clear). Our algorithm also bears some similarity to one used in Van Roy [12] for value determination in the setting of fully observable MDPs.

The remainder of this paper is structured as follows: Section 2 defines the notation that will be used in this paper, and formalizes the concepts of deterministic simulative models and of families of realizable dynamics. Section 3 then describes how we transform POMDPs into ones with only deterministic transitions, and gives our policy search algorithm. Section 4 goes on to establish conditions under which we may give guarantees on the performance of the algorithm, Section 5 describes our experimental results, and Section 6 closes with conclusions.

## 2　Preliminaries

This section gives our notation, and introduces the concept of the set of realizable dynamics of a POMDP under a policy class.

A Markov decision process (MDP) is a tuple $(S, D, A, \{P_{sa}(\cdot)\}, \gamma, R)$ where: $S$ is a set of **states**; $D$ is the **initial-state distribution**, from which the start-state $s_0$ is drawn; $A$ is a set of **actions**; $\{P_{sa}(\cdot)\}$ are the **transition probabilities**, with $P_{sa}$ giving the next-state distribution upon taking action $a$ in state $s$; $\gamma \in [0,1)$ is the **discount factor**; and $R$ is the **reward function**, bounded by $R_{\max}$. For the sake of concreteness, we will assume, unless otherwise stated, that $S = [0,1]^{d_S}$ is a $d_S$-dimensional hypercube. For simplicity, we also assume rewards are deterministic, and written $R(s)$ rather than $R(s,a)$, the extensions being trivial. Lastly, everything that needs to be measurable is assumed to be measurable.

A policy is any mapping $\pi : S \mapsto A$. The **value function** of a policy $\pi$ is a map $V^\pi : S \mapsto \mathbb{R}$, so that $V^\pi(s)$ gives the expected discounted sum of rewards for executing $\pi$ starting from state $s$. With some abuse of notation, we also define the value of a *policy*, with respect to the initial-state distribution $D$, according to

$$V(\pi) = \mathbf{E}_{s_0 \sim D}[V^\pi(s_0)] \qquad (1)$$

(where the subscript $s_0 \sim D$ indicates that the expectation is with respect to $s_0$ drawn according to $D$). When we are considering multiple MDPs and wish to make explicit that a value function is for a particular MDP $M$, we will also write $V_M^\pi(s)$, $V_M(\pi)$, etc.

In the policy search setting, we have some fixed class $\Pi$ of policies, and desire to find a good policy $\pi \in \Pi$. More precisely, for a given MDP $M$ and policy class $\Pi$, define

$$opt(M, \Pi) = \sup_{\pi \in \Pi} V_M(\pi). \qquad (2)$$

Our goal is to find a policy $\hat{\pi} \in \Pi$ so that $V(\hat{\pi})$ is close to $opt(M, \Pi)$.

Note that this framework also encompasses cases where our family $\Pi$ consists of policies that depend only on certain aspects of the state. In particular, in POMDPs, we can restrict attention to policies that depend only on the observables. This restriction results in a subclass of stochastic memory-free policies.[1] By introducing artificial "memory variables" into the process state, we can also define stochastic limited-memory policies [9] (which certainly permits some *belief state* tracking).

---

[1] Although we have not explicitly addressed stochastic policies so far, they are a straightforward generalization (e.g. using the transformation to deterministic policies given in [7]).

Since we are interested in the "planning" problem, we assume that we are given a model of the (PO)MDP. Much previous work has studied the case of (PO)MDPs specified via a generative model [7, 13], which is a stochastic function that takes as input any $(s,a)$ state-action pair, and outputs $s'$ according to $P_{sa}(\cdot)$ (and the associated reward). In this paper, we assume a stronger model. We assume we have a *deterministic* function $g : S \times A \times [0,1]^{d_P} \mapsto S$, so that for any fixed $(s,a)$-pair, if $\vec{p}$ is distributed Uniform$[0,1]^{d_P}$, then $g(s, a, \vec{p})$ is distributed according to the transition distribution $P_{sa}(\cdot)$. In other words, to draw a sample from $P_{sa}(\cdot)$ for some fixed $s$ and $a$, we need only draw $\vec{p}$ uniformly in $[0,1]^{d_P}$, and then take $g(s, a, \vec{p})$ to be our sample. We will call such a model a deterministic simulative model for a (PO)MDP.

Since a deterministic simulative model allows us to simulate a generative model, it is clearly a stronger model. However, most computer implementations of generative models also provide deterministic simulative models. Consider a generative model that is implemented via a procedure that takes $s$ and $a$, makes at most $d_P$ calls to a random number generator, and then outputs $s'$ drawn according to $P_{sa}(\cdot)$. Then this procedure is already providing a deterministic simulative model. The only difference is that the deterministic simulative model has to make explicit (or "expose") its interface to the random number generator, via $\vec{p}$. (A generative model implemented via a physical simulation of an MDP with "resets" to arbitrary states does not, however, readily lead to a deterministic simulative model.)

Let us examine some simple examples of deterministic simulative models. Suppose that for a state-action pair $(s_1, a_1)$ and some states $s'$ and $s''$, $P_{s_1 a_1}(s') = 1/3$, $P_{s_1 a_1}(s'') = 2/3$. Then we may choose $d_P = 1$ so that $\vec{p} = p$ is just a real number, and let $g(s_1, a_1, p) = s'$ if $p \leq 1/3$, and $g(s_1, a_1, p) = s''$ otherwise. As another example, suppose $S = \mathbb{R}$, and $P_{sa}(\cdot)$ is a normal distribution with a cumulative distribution function $F_{sa}(\cdot)$. Again letting $d_P = 1$, we may choose $g$ to be $g(s, a, p) = F_{sa}^{-1}(p)$.

It is a fact of probability and measure theory that, given any transition distribution $P_{sa}(\cdot)$, such a deterministic simulative model $g$ can always be constructed for it. (See, e.g. [4].) Indeed, some texts (e.g. [2]) routinely define POMDPs using essentially deterministic simulative models. However, there will often be many different choices of $g$ for representing a (PO)MDP, and it will be up to the user to decide which one is most "natural" to implement. As we will see later, the particular choice of $g$ that the user makes can indeed impact the performance of our algorithm, and "simpler" (in a sense to be formalized) implementations are generally preferred.

To close this section, we introduce a concept that will be useful later, that captures the family of dynamics that a (PO)MDP and policy class can exhibit. Assume a deterministic simulative model $g$, and fix a policy $\pi$. If we are executing $\pi$ from some state $s$, the successor-state is determined by $f_\pi(s, \vec{p}) = g(s, \pi(s), \vec{p})$, which is a function of $s$ and $\vec{p}$. Varying $\pi$ over $\Pi$, we get a whole family of functions $\mathcal{F} = \{f_\pi | f_\pi(s, \vec{p}) = g(s, \pi(s), \vec{p})\}$ mapping from $S \times [0, 1]^{d_P}$ into successor states $S$. This set of functions $\mathcal{F}$ should be thought of as the family of dynamics realizable by the POMDP and $\Pi$, though since its definition does depend on the particular deterministic simulative model $g$ that we have chosen, this is "as expressed with respect to $g$." For each $f$, also let $f_i$ be the $i$-th coordinate function (so that $f_i(s, \vec{p})$ is the $i$-th coordinate of $f(s, \vec{p})$) and let $\mathcal{F}_i$ be the corresponding families of coordinate functions mapping from $S \times [0, 1]^{d_P}$ into $[0, 1]$. Thus, $\mathcal{F}_i$ captures all the ways that coordinate $i$ of the state can evolve.

We are now ready to describe our policy search method.

## 3 Policy search method

In this section, we show how we transform a (PO)MDP into an "equivalent" one that has only deterministic transitions. This then leads to natural estimates $\hat{V}(\pi)$ of the policies' values $V(\pi)$. Finally, we may search over policies to optimize $\hat{V}(\pi)$, to find a (hopefully) good policy.

### 3.1 Transformation of (PO)MDPs

Given a (PO)MDP $M = (S, D, A, \{P_{sa}(\cdot)\}, \gamma, R)$ and a policy class $\Pi$, we describe how, using a deterministic simulative model $g$ for $M$, we construct our transformed POMDP $M' = (S', D', A, \{P'_{sa}(\cdot)\}, \gamma, R')$ and corresponding class of policies $\Pi'$, so that $M'$ has only deterministic transitions (though its initial state may still be random). To simplify the exposition, we assume $d_P = 1$, so that the terms $\vec{p}$ are just real numbers.

$M'$ is constructed is as follows: The action space and discount factor for $M'$ are the same as in $M$. The state space for $M'$ is $S \times [0, 1]^\infty$. In other words, a typical state in $M'$ can be written as a vector $(s, p_1, p_2, \ldots)$ — this consists of a state $s$ from the original state space $S$, followed by an infinite sequence of real numbers in $[0, 1]$.

The rest of the transformation is straightforward. Upon taking action $a$ in state $(s, p_1, p_2, \ldots)$ in $M'$, we *deterministically* transition to the state $(s', p_2, p_3, \ldots)$, where $s' = g(s, a, p_1)$. In other words, the $s$ portion of the state (which should be thought of as the "actual" state) changes to $s'$, and one number in the infinite sequence $(p_1, p_2, \ldots)$ is used up to generate $s'$ from the correct distribution. By the definition of the deterministic simulative model $g$, we see that so long as $p_1 \sim$ Uniform$[0, 1]$, then the "next-state" distribution of $s'$ is the same as if we had taken action $a$ in state $s$ (randomization over $p_1$).

Finally, we choose $D'$, the initial-state distribution over $S' = S \times [0, 1]^\infty$, so that $(s, p_1, p_2, \ldots)$ drawn according to $D'$ will be so that $s \sim D$, and the $p_i$'s are distributed i.i.d. Uniform$[0, 1]$. For each policy $\pi \in \Pi$, also let there be a corresponding $\pi' \in \Pi'$, given by $\pi'(s, p_1, p_2, \ldots) = \pi(s)$, and let the reward be given by $R'(s, p_1, p_2, \ldots) = R(s)$.

If one observes only the "$s$"-portion (but not the $p_i$'s) of a sequence of states generated in the POMDP $M'$ using policy $\pi'$, one obtains a sequence that is drawn from the same distribution as would have been generated from the original (PO)MDP $M$ under the corresponding policy $\pi \in \Pi$. It follows that, for corresponding policies $\pi \in \Pi$ and $\pi' \in \Pi'$, we have that $V_M(\pi) = V_{M'}(\pi')$. This also implies that the best possible expected returns in both (PO)MDPs are the same: $opt(M, \Pi) = opt(M', \Pi')$.

To summarize, we have shown how, using a deterministic simulative model, we can transform any POMDP $M$ and policy class $\Pi$ into an "equivalent" POMDP $M'$ and policy class $\Pi'$, so that the transitions in $M'$ are deterministic; i.e., given a state $s \in S'$ and an action $a \in A$, the next-state in $M'$ is exactly determined. Since policies in $\Pi$ and $\Pi'$ have the same values, if we can find a policy $\pi' \in \Pi'$ that does well in $M'$ starting from $D'$, then the corresponding policy $\pi \in \Pi$ will also do well for the original POMDP $M$ starting from $D$. Hence, the problem of policy search in general POMDPs is reduced to the problem of policy search in POMDPs with *deterministic* transition dynamics. In the next section, we show how we can exploit this fact to derive a simple and natural policy search method.

### 3.2 PEGASUS: A method for policy search

As discussed, it suffices for policy search to find a good policy $\pi' \in \Pi'$ for the transformed POMDP, since the corresponding policy $\pi \in \Pi$ will be just as good. To do this, we first construct an approximation $\hat{V}_{M'}(\cdot)$ to $V_M(\cdot)$, and then search over policies $\pi' \in \Pi'$ to optimize $\hat{V}_{M'}(\pi')$ (as a proxy for optimizing the hard-to-compute $V_M(\pi)$), and thus find a (hopefully) good policy.

Recall that $V_{M'}$ is given by

$$V_{M'}(\pi) = \mathbf{E}_{s_0 \sim D'} [V^\pi_{M'}(s_0)], \qquad (3)$$

where the expectation is over the initial state $s_0 \in S'$ drawn according to $D'$. The first step in the approximation is to replace the expectation over the distribution with a finite sample of states. More precisely, we first draw a sample $\{s_0^{(1)}, s_0^{(2)}, \ldots, s_0^{(m)}\}$ of $m$ initial states according to $D'$. These states, also called "scenarios" (a term from the stochastic optimization literature; see, e.g. [3]), define an approximation to $V_{M'}(\pi)$:

$$V_{M'}(\pi) \approx \frac{1}{m} \sum_{i=1}^{m} V^\pi_{M'}(s_0^{(i)}). \qquad (4)$$

Since the transitions in $M'$ are deterministic, for a given state $s \in S'$ and a policy $\pi \in \Pi'$, the sequence of states that will be visited upon executing $\pi$ from $s$ is exactly determined; hence the sum of discounted rewards for executing $\pi$ from $s$ is also exactly determined. Thus, to calculate one of the terms $V_{M'}^\pi(s_0^{(i)})$ in the summation in Equation (4) corresponding to scenario $s_0^{(i)}$, we need only use our deterministic simulative model to find the sequence of states visited by executing $\pi$ from $s_0^{(i)}$, and sum up the resulting discounted rewards. Naturally, this would be an infinite sum, so the second (and standard) part of the approximation is to truncate this sum after some number $H$ of steps, where $H$ is called the horizon time. Here, we choose $H$ to be the $\epsilon$-horizon time $H_\epsilon = \log_\gamma(\epsilon(1-\gamma)/2R_{\max})$, so that (because of discounting) the truncation introduces at most $\epsilon/2$ error into the approximation.

To summarize, given $m$ scenarios $s_0^{(1)}, \ldots, s_0^{(m)}$, our approximation to $V_{M'}$ is the deterministic function

$$\hat{V}_{M'}(\pi) = \frac{1}{m} \sum_{i=1}^m R'(s_0^{(i)}) + \gamma R'(s_1^{(i)}) + \cdots + \gamma^{H_\epsilon} R'(s_{H_\epsilon}^{(i)})$$

where $(s_0^{(i)}, s_1^{(i)}, \ldots, s_{H_\epsilon}^{(i)})$ is the sequence of states deterministically visited by $\pi$ starting from $s_0^{(i)}$. Given $m$ scenarios, this defines an approximation to $V_{M'}(\pi)$ for *all* policies $\pi \in \Pi'$.

The final implementation detail is that, since the states $s_0^{(i)} \in S \times [0,1]^\infty$ are infinite-dimensional vectors, we have no way of representing them (and their successor states) explicitly. But because we will be simulating only $H_\epsilon$ steps, we need only represent $p_1^{(i)}, p_2^{(i)}, \ldots, p_{H_\epsilon}^{(i)}$, of the state $s_0^{(i)} = (s^{(i)}, p_1^{(i)}, p_2^{(i)}, \ldots)$, and so we will do just that. Viewed in the space of the original, untransformed POMDP, evaluating a policy this way is therefore also akin to generating $m$ Monte Carlo trajectories and taking their empirical average return, but with the crucial difference that all the randomization is "fixed" in advance and "reused" for evaluating different $\pi$.

Having used $m$ scenarios to define $\hat{V}_{M'}(\pi)$ for all $\pi$, we may search over policies to optimize $\hat{V}_{M'}(\pi)$. We call this policy search method PEGASUS: Policy Evaluation-of-Goodness And Search Using Scenarios. Since $\hat{V}_{M'}(\pi)$ is a deterministic function, the search procedure only needs to optimize a deterministic function, and any number of standard optimization methods may be used. In the case that the action space is continuous and $\Pi = \{\pi_\theta | \theta \in \mathbb{R}^\ell\}$ is a smoothly parameterized family of policies (so $\pi_\theta(s)$ is differentiable in $\theta$ for all $s$) then if all the relevant quantities are differentiable, it is also possible to find the derivatives $(d/d\theta)\hat{V}_{M'}(\pi_\theta)$, and gradient ascent methods can be used to optimize $\hat{V}_{M'}(\pi_\theta)$. One common barrier to doing this is that $R$ is often discontinuous, being (say) 1 within a goal region and 0 elsewhere. One approach to dealing with this problem is to smooth $R$ out, possibly in combination with "continuation" methods that gradually unsmooth it again. An alternative approach that may be useful in the setting of continuous dynamical systems is to alter the reward function to use a continuous-time model of discounting. Assuming that the time at which the agent enters the goal region is differentiable, then $\hat{V}_{M'}(\pi_\theta)$ is again differentiable.[2]

## 4 Main theoretical results

PEGASUS samples a number of scenarios from $D'$, and uses them to form an approximation $\hat{V}(\pi)$ to $V(\pi)$. If $\hat{V}$ is a uniformly good approximation to $V$, then we can guarantee that optimizing $\hat{V}$ will result in a policy with value close to $opt(M, \Pi)$. This section establishes conditions under which this occurs.

### 4.1 The case of finite action spaces

We begin by considering the case of two actions, $A = \{a_1, a_2\}$. Studying policy search in a similar setting, Kearns, Mansour and Ng [7] established conditions under which their algorithm gives uniformly good estimates of the values of policies. A key to that result was that uniform convergence can be established so long as the policy class $\Pi$ has low "complexity." This is analogous to the setting of supervised learning, where a learning algorithm that uses a hypothesis class $\mathcal{H}$ that has low complexity (such as in the sense of low VC-dimension) will also enjoy uniform convergence of its error estimates to their means.

In our setting, since $\Pi$ is just a class of functions mapping from $S$ into $\{a_1, a_2\}$, it is just a set of boolean functions. Hence, $\text{VC}(\Pi)$, its Vapnik-Chervonenkis dimension [14], is well defined. That is, we say $\Pi$ *shatters* a set of $m$ states if it can realize each of the $2^m$ possible action combinations on them, and $\text{VC}(\Pi)$ is just the size of the largest set shattered by $\Pi$. The result of Kearns et al. then suffices to give the following theorem.[3]

**Theorem 1** *Let a POMDP with actions $A = \{a_1, a_2\}$ be given, and let $\Pi$ be a class of strategies for this POMDP, with Vapnik-Chervonenkis dimension $d = \text{VC}(\Pi)$. Also let any $\epsilon, \delta > 0$ be fixed, and let $\hat{V}$ be the policy-value estimates determined by* PEGASUS *using $m$ scenarios and*

---

[2]More precisely, if the agent enters the goal region on some time step, then rather than giving it a reward of 1, we figure out what fraction $\tau \in [0,1]$ of that time step (measured in continuous time) the agent had taken to enter the goal region, and then give it reward $\gamma^\tau$ instead. Assuming $\tau$ is differentiable in the system's dynamics, then $\gamma^\tau$ and hence $\hat{V}_{M'}(\pi_\theta)$ are now also differentiable (other than on a usually-measure 0 set, for example from truncation at $H_\epsilon$ steps).

[3]The algorithm of Kearns, Mansour and Ng uses a "trajectory tree" method to find the estimates $\hat{V}(\pi)$; since each trajectory tree is of size $\exp(O(H_\epsilon))$, they were very expensive to build. Each scenario in PEGASUS can be viewed as a compact representation of a trajectory tree (with a technical difference that different subtrees are not constructed independently), and the proof given in Kearns et al. then applies without modification to give Theorem 1.

*a horizon time of* $H_\epsilon$. *If*

$$m = O\left(\text{poly}\left(d, \frac{R_{\max}}{\epsilon}, \log\frac{1}{\delta}, \frac{1}{1-\gamma}\right)\right), \quad (5)$$

*then with probability at least* $1 - \delta$, $\hat{V}$ *will be uniformly close to* $V$:

$$\left|\hat{V}(\pi) - V(\pi)\right| \leq \epsilon \quad \text{for all } \pi \in \Pi \quad (6)$$

Using the transformation given in Kearns et al., the case of a finite action space with $|A| > 2$ also gives rise to essentially the same uniform-convergence result, so long as $\Pi$ has low "complexity."

The bound given in the theorem has *no dependence* on the size of the state space or on the "complexity" of the POMDP's transitions and rewards. Thus, so long as $\Pi$ has low VC-dimension, uniform convergence will occur, independently of how complicated the POMDP is. As in Kearns et al., this theorem therefore recovers the best analogous results in supervised learning, in which uniform convergence occurs so long as the hypothesis class has low VC-dimension, regardless of the size or "complexity" of the underlying space and target function.

### 4.2 The case of infinite action spaces: "Simple" $\Pi$ is insufficient for uniform convergence

We now consider the case of infinite action spaces. Whereas, in the 2-action case, $\Pi$ being "simple" was sufficient to ensure uniform convergence, this is not the case in POMDPs with infinite action spaces.

Suppose $A$ is a (countably or uncountably) infinite set of actions. A "simple" class of policies would be $\Pi = \{\pi_a | \pi_a(s) \equiv a, a \in A\}$ — the set of all policies that always choose the same action, regardless of the state. Intuitively, this is the simplest policy that actually uses an infinite action space; also, any reasonable notion of complexity of policy classes should assign $\Pi$ a low "dimension." If it were true that simple policy classes imply uniform convergence, then it is certainly true that this $\Pi$ should always enjoy uniform convergence. Unfortunately, this is not the case, as we now show.

**Theorem 2** *Let $A$ be an infinite set of actions, and let $\Pi = \{\pi_a | \pi_a(s) \equiv a, a \in A\}$ be the corresponding set of all "constant valued" policies. Then there exists a finite-state MDP with action space $A$, and a deterministic simulative model for it, so that* PEGASUS' *estimates using the deterministic simulative model do not uniformly converge to their means. i.e. There is an $\epsilon > 0$, so that for estimates $\hat{V}$ derived using any finite number $m$ of scenarios and any finite horizon time, there is a policy $\pi \in \Pi$ so that*

$$|\hat{V}(\pi) - V(\pi)| > \epsilon. \quad (7)$$

The proof of this Theorem, which is not difficult, is in Appendix A. This result shows that simplicity of $\Pi$ is not sufficient for uniform convergence in the case of infinite action spaces. However, the counterexample used in the proof of Theorem 2 has a very complex $g$ despite the MDP being quite simple. Indeed, a different choice for $g$ would have made uniform convergence occur.[4] Thus, it is natural to hypothesize that assumptions on the "complexity" of $g$ are also needed to ensure uniform convergence. As we will shortly see, this intuition is roughly correct. Since actions affect transitions only through $g$, the crucial quantity is actually the composition of policies and the deterministic simulative model — in other words, the class $\mathcal{F}$ of the dynamics realizable in the POMDP and policy class, using a particular deterministic simulative model. In the next section, we show how assumptions on the complexity of $\mathcal{F}$ leads to uniform convergence bounds of the type we desire.

### 4.3 Uniform convergence in the case of infinite action spaces

For the remainder of this section, assume $S = [0,1]^{d_S}$. Then $\mathcal{F}$ is a class of functions mapping from $[0,1]^{d_S} \times [0,1]^{d_P}$ into $[0,1]^{d_S}$, and so a simple way to capture its "complexity" is to capture the complexity of its families of coordinate functions, $\mathcal{F}_i$, $i = 1, \ldots, d_S$. Each $\mathcal{F}_i$ is a family of functions mapping from $[0,1]^{d_S} \times [0,1]^{d_P}$ into $[0,1]$, the $i$-th coordinate of the state vector. Thus, $\mathcal{F}_i$ is just a family of real-valued functions — the family of $i$-th coordinate dynamics that $\Pi$ can realize, with respect to $g$.

The complexity of a class of boolean functions is measured by its VC dimension, defined to be the size of the largest set shattered by the class. To capture the "complexity" of real-valued families of functions such as $\mathcal{F}_i$, we need a generalization of the VC dimension. The pseudo-dimension, due to Pollard [10] is defined as follows:

**Definition (Pollard, 1990).** Let $\mathcal{H}$ be a family of functions mapping from a space $X$ into $\mathbb{R}$. Let a sequence of $d$ points $x_1, \ldots, x_d \in X$ be given. We say $\mathcal{H}$ *shatters* $x_1, \ldots, x_d$ if there exists a sequence of real numbers $t_1, \ldots, t_d$ such that the subset of $\mathbb{R}^d$ given by $\{(h(x_1) - t_1, \ldots, h(x_d) - t_d) | h \in \mathcal{H}\}$ intersects all $2^d$ orthants of $\mathbb{R}^d$ (equivalently, if for any sequence of $d$ bits $b_1, \ldots, b_d \in \{0, 1\}$, there is a function $h \in \mathcal{H}$ such that $h(x_i) \geq t_i \Leftrightarrow b_i = 1$, for all $i = 1, \ldots, d$). The **pseudo-dimension** of $\mathcal{H}$, denoted $\dim_P(\mathcal{H})$, is the size of the largest set that $\mathcal{H}$ shatters, or infinite if $\mathcal{H}$ can shatter arbitrarily large sets.

The pseudo-dimension generalizes the VC dimension, and coincides with it in the case that $\mathcal{H}$ maps into $\{0, 1\}$. We will use it to capture the "complexity" of the classes of the POMDP's realizable dynamics $\mathcal{F}_i$. We also remind readers of the definition of Lipschitz continuity.

---

[4]For example, $g(s_0, a, p) = s_{-1}$ if $p \leq 0.5$, $s_1$ otherwise; see Appendix A.

**Definition.** A function $f : \mathbb{R}^n \mapsto \mathbb{R}$ is **Lipschitz continuous** (with respect to the Euclidean norm on its range and domain) if there exists a constant $B$ such that for all $x, y \in \text{dom}(f)$, $||f(x) - f(y)||_2 \leq B||x - y||_2$. Here, $B$ is called a **Lipschitz bound**. A family of functions $\mathcal{H}$ mapping from $\mathbb{R}^n$ into $\mathbb{R}$ is **uniformly Lipschitz continuous** with Lipschitz bound $B$ if every function $h \in \mathcal{H}$ is Lipschitz continuous with Lipschitz bound $B$.

We now state our main theorem, with a corollary regarding when optimizing $\hat{V}$ will result in a provably good policy.

**Theorem 3** *Let a POMDP with state space $S = [0, 1]^{d_S}$, and a possibly infinite action space be given. Also let a policy class $\Pi$, and a deterministic simulative model $g : S \times A \times [0, 1]^{d_P} \mapsto S$ for the POMDP be given. Let $\mathcal{F}$ be the corresponding family of realizable dynamics in the POMDP, and $\mathcal{F}_i$ the resulting families of coordinate functions. Suppose that $\dim_P(\mathcal{F}_i) \leq d$ for each $i = 1, \ldots, d_S$, and that each family $\mathcal{F}_i$ is uniformly Lipschitz continuous with Lipschitz bound at most $B$, and that the reward function $R : S \mapsto [-R_{\max}, R_{\max}]$ is also Lipschitz continuous with Lipschitz bound at most $B_R$. Finally, let $\epsilon, \delta > 0$ be given, and let $\hat{V}$ be the policy-value estimates determined by PEGASUS using $m$ scenarios and a horizon time of $H_\epsilon$. If $m =$*

$$O\left(\text{poly}\left(d, \frac{R_{\max}}{\epsilon}, \log\frac{1}{\delta}, \frac{1}{1-\gamma}, \log B, \log\frac{B_R}{R_{\max}}, d_S, d_P\right)\right)$$

*then with probability at least $1 - \delta$, $\hat{V}$ will be uniformly close to $V$:*

$$\left|\hat{V}(\pi) - V(\pi)\right| \leq \epsilon \quad \text{for all } \pi \in \Pi \quad (8)$$

**Corollary 4** *Under the conditions of Theorem 1 or 3, let $m$ be chosen as in the Theorem. Then with probability at least $1 - \delta$, the policy $\hat{\pi}$ chosen by optimizing the value estimates, given by $\hat{\pi} = \arg\max_{\pi \in \Pi} \hat{V}(\pi)$, will be near-optimal in $\Pi$:*

$$V(\hat{\pi}) \geq \text{opt}(M, \Pi) - 2\epsilon \quad (9)$$

**Remark.** The (Lipschitz) continuity assumptions give a sufficient but not necessary set of conditions for the theorem, and other sets of sufficient conditions can be envisaged. For example, if we assume that the distribution on states induced by any policy at each time step has a bounded density, then we can show uniform convergence for a large class of ("reasonable") discontinuous reward functions such as $R(s) = 1$ if $s_1 > 0.5$, 0 otherwise.[5]

---

[5] Space constraints preclude a detailed discussion, but briefly, this is done by constructing two Lipschitz continuous reward functions $R_U$ and $R_L$ that are "close to" and which upper- and lower-bound $R$ (and which hence give value estimates that also upper- and lower-bound our value estimates under $R$); using the assumption of bounded densities to show our values under $R_U$ and $R_L$ are $\epsilon$-close to that of $R$; applying Theorem 3 to show uniform convergence occurs with $R_U$ and $R_L$; and lastly deducing from this that uniform convergence occurs with $R$ as well.

Using tools from [5], it is also possible to show similar uniform convergence results without Lipschitz continuity assumptions, by assuming that the family $\pi$ is parameterized by a small number of real numbers, and that $\pi$ (for all $\pi \in \Pi$), $g$, and $R$ are each implemented by a function that calculates their results using only a bounded number of the usual arithmetic operations on real numbers.

The proof of Theorem 3, which uses techniques first introduced by Haussler [6] and Pollard [10], is quite lengthy, and is deferred to Appendix B.

## 5 Experiments

In this section, we report the results from two experiments. The first, run to examine the behavior of PEGASUS parametrically, involved a simple gridworld POMDP. The second studied a complex continuous state/continuous action problem involving riding a bicycle.

Figure 1a shows the finite state and action POMDP used in our first experiment. In this problem, the agent starts in the lower-left corner, and receives a $-1$ reinforcement per step until it reaches the absorbing state in the upper-right corner. The eight possible observations, also shown in the figure, indicate whether each of the eight squares adjoining the current position contains a wall. The policy class is small, consisting of all $4^8 = 65536$ functions mapping from the eight possible observations to the four actions corresponding to trying to move in each of the compass directions. Actions are noisy, and result in moving in a random direction 20% of the time. Since the policy class is small enough to exhaustively enumerate, our optimization algorithm for searching over policies was simply exhaustive search, trying all $4^8$ policies on the $m$ scenarios, and picking the best one. Our experiments were done with $\gamma = 0.99$ and a horizon time of $H = 100$, and all results reported on this problem are averages over 10000 trials. The deterministic simulative model was

$$g(s, a, p) = \begin{cases} \delta(s, \text{up}) & \text{if } p \leq 0.05 \\ \delta(s, \text{left}) & \text{if } 0.05 < p \leq 0.10 \\ \delta(s, \text{down}) & \text{if } 0.10 < p \leq 0.15 \\ \delta(s, \text{right}) & \text{if } 0.15 < p \leq 0.20 \\ \delta(s, a) & \text{otherwise} \end{cases}$$

where $\delta(s, a)$ denotes the result of moving one step from $s$ in the direction indicated by $a$, and is $s$ if this move would result in running into a wall.

Figure 1b shows the result of running this experiment, for different numbers of scenarios. The value of the best policy within $\Pi$ is indicated by the topmost horizontal line, and the solid curve below that is the mean policy value when using our algorithm. As we see, even using surprisingly small numbers of scenarios, the algorithm manages to find good policies, and as $m$ becomes large, the value also approaches the optimal value.

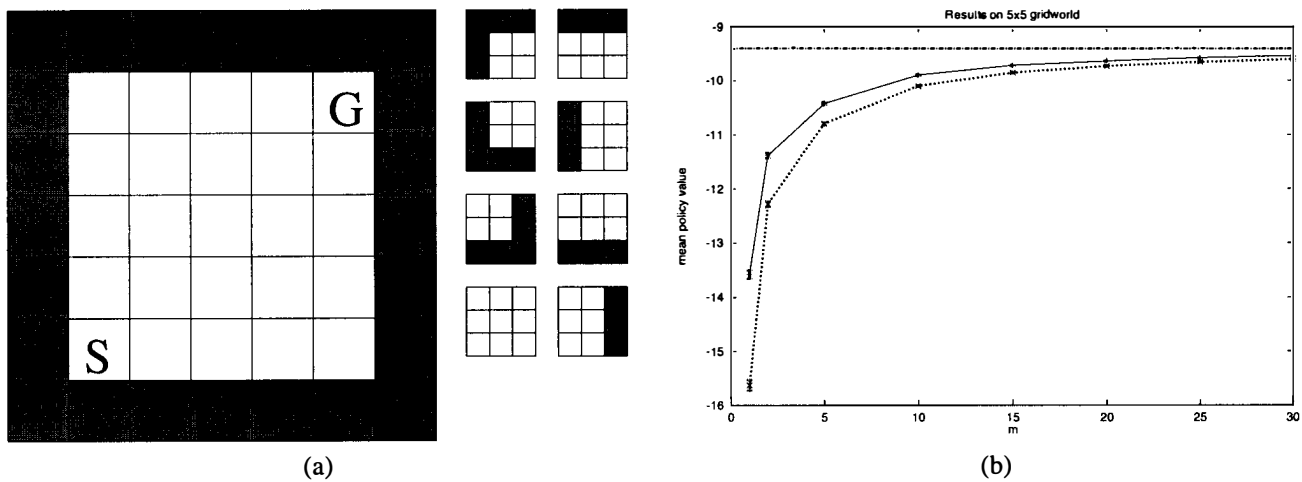

Figure 1: (a) 5x5 gridworld, with the 8 observations. (b) PEGASUS results using the normal and complex deterministic simulative models. The topmost horizontal line shows the value of the best policy in $\Pi$; the solid curve is the mean policy value using the normal model; the lower curve is the mean policy value using the complex model. The (almost negligible) 1 s.e. bars are also plotted.

We had previously predicted that a "complicated" deterministic simulative model $g$ can lead to poor results. For each $(s, a)$-pair, let $h_{s,a} : [0, 1] \mapsto [0, 1]$ be a hash function that maps any Uniform[0, 1] random variable into another Uniform[0, 1] random variable.[6] Then if $g$ is a deterministic simulative model, $g'(s, a, p) = g(s, a, h_{s,a}(p))$ is another one that, because of the presence of the hash function, is a much more "complex" model than $g$. (Here, we appeal to the reader's intuition about complex functions, rather than formal measures of complexity.) We would therefore predict that using PEGASUS with $g'$ would give worse results than $g$, and indeed this prediction is borne out by the results as shown in Figure 1b (dashed curve). The difference between the curves is not large, and this is also not unexpected given the small size of the problem.[7]

Our second experiment used Randløv and Alstrøm's [11] bicycle simulator, where the objective is to ride to a goal one kilometer away. The actions are the torque $\tau$ applied to the handlebars and the displacement $\nu$ of the rider's center-of-gravity from the center. The six-dimensional state used in [11] includes variables for the bicycle's tilt angle and orientation, and the handlebar's angle. If the bicycle tilt exceeds $\pi/15$, it falls over and enters an absorbing state, receiving a large negative reward. The randomness in the simulator is from a uniformly distributed term added to the intended displacement of the center-of-gravity. Rescaled appropriately, this became the $p$ term of our deterministic simulative model.

We performed policy search over the following space: We selected a vector $\vec{x}$ of fifteen (simple, manually-chosen but not fine-tuned) features of each state; actions were then chosen with sigmoids: $\tau = \sigma(w_1 \cdot \vec{x})(\tau_{\max} - \tau_{\min}) + \tau_{\min}$, $\nu = \sigma(w_2 \cdot \vec{x})(\nu_{\max} - \nu_{\min}) + \nu_{\min}$, where $\sigma(z) = 1/(1 + e^{-z})$. Note that since our approach can handle continuous actions directly, we did not, unlike [11], have to discretize the actions. The initial-state distribution was manually chosen to be representative of a "typical" state distribution when riding a bicycle, and was also not fine-tuned. We used only a small number $m = 30$ of scenarios, $\gamma = 0.998$, $H = 500$, with the continuous-time model of discounting discussed earlier, and (essentially) gradient ascent to optimize over the weights.[8] Shaping rewards, to reward progress towards the goal, were also used.[9]

We ran 10 trials using our policy search algorithm, testing each of the resulting solutions on 50 rides. Doing so, the median riding distances to the goal of the 10 different policies ranged from about 0.995km[10] to 1.07km. In all 500 evaluation runs for the 10 policies, the worst distance we observed was also about 1.07km. These results are *significantly* better than those of [11], which reported riding distances of about 7km (since their policies often took very "non-linear" paths to the goal), and a single "best-ever" trial of about 1.7km.

---

[6]In our experiments, this was implemented by choosing, for each $(s, a)$ pair, a random integer $k(s, a)$ from $\{1, \ldots, 1000\}$, and then letting $h_{s,a}(p) = \text{fract}(k(s, a) \cdot p)$, where $\text{fract}(x)$ denotes the fractional part of $x$.

[7]Theory predicts that the difference between $g$ and $g'$'s performance should be at most $O(\sqrt{\log |\Pi|/m})$; see [7].

[8]Running experiments without the continuous-time model of discounting, we also obtained, using a non-gradient based hill-climbing algorithm, equally good results as those reported here. Our implementation of gradient ascent, using numerically evaluated derivates, was run with a bound on the length of a step taken on any iteration, to avoid problems near $\hat{V}(\pi_\theta)$'s discontinuities.

[9]Other experimental details: The shaping reward was proportional to and signed the same as the amount of progress towards the goal. As in [11], we did not include the distance-from-goal as one of the state variables during training; training therefore proceeding "infinitely distant" from the goal.

[10]Distances under 1km are possible since, as in [11], the goal has a 10m radius.

## 6 Conclusions

We have shown how any POMDP can be transformed into an "equivalent" one in which all transitions are deterministic. By approximating the transformed POMDP's initial state distribution with a sample of scenarios, we defined an estimate for the value of every policy, and finally performed policy search by optimizing these estimates. Conditions were established under which these estimates will be uniformly good, and experimental results showed our method working well. It is also straightforward to extend these methods and results to the cases of finite-horizon undiscounted reward, and infinite-horizon average reward with $\epsilon$-mixing time $H_\epsilon$.

**Acknowledgements**

We thank Jette Randløv and Preben Alstrøm for the use of their bicycle simulator, and Ben Van Roy for helpful comments. A. Ng is supported by a Berkeley Fellowship. This work was also supported by ARO MURI DAAH04-96-0341, ONR MURI N00014-00-1-0637, and NSF grant IIS-9988642.

## Appendix A: Proof of Theorem 2

**Proof (of Theorem 2).** We construct an MDP with states $s_{-1}, s_0$, and $s_1$ plus an absorbing state. The reward function is $R(s_i) = i$ for $i = -1, 0, 1$. Discounting is ignored in this construction. Both $s_{-1}$ and $s_1$ transition with probability 1 to the absorbing state regardless of the action taken. The initial-state $s_0$ has a .5 chance of transitioning to each of $s_{-1}$ and $s_1$.

We now construct $g$, which will depend in a complicated way on the $\vec{p}$ term. Let $U = \{\cup_{i=1}^N [a_i, b_i] \,|\, a_i, b_i \in [0,1] \cap \mathbb{Q}, a_i < b_i, 1 \leq N < \infty\}$ be the countable set of all finite unions of intervals with rational endpoints in $[0,1]$. Let $U'$ be the countable subset of $U$ that contains all elements of $U$ that have total length (Lebesgue measure) exactly 0.5. For example, $[1/3, 5/6]$ and $[0.0, 0.25] \cup [0.5, 0.75]$ are both in $U'$. Let $U_1, U_2, \ldots$ be an enumeration of the elements of $U'$. Also let $\{a_1, a_2, \ldots\}$ be an enumeration of (some countably infinite subset of) $A$. The deterministic simulative model on these actions is given by:

$$g(s_0, a_i, p) = \begin{cases} s_{-1} & \text{if } p \in U_i \\ s_1 & \text{otherwise} \end{cases}$$

So, $P_{s_0 a_i}(s_1) = P_{s_0 a_i}(s_{-1}) = 0.5$ for all $a_i$, and this is a correct model for the MDP. Note also that $V(\pi) = 0$ for all $\pi \in \Pi$.

For any finite sample of $m$ scenarios $(s_0, \vec{p}^{(1)}), (s_0, \vec{p}^{(2)}), \ldots, (s_0, \vec{p}^{(m)})$, there exists some $U_i$ such that $p_1^{(j)} \notin U_i$ for all $j = 1, \ldots, m$. Thus, evaluating $\pi_i \equiv a_i$ using this set of scenarios, all $m$ simulated trajectories will transition from $s_0$ from $s_1$, so the value estimate (assuming $H_\epsilon \geq 1$) for $\pi_i$ is $\hat{V}(\pi_i) = 1$. Since this argument holds for any finite number $m$ of scenarios, we have shown that $\hat{V}$ does not uniformly converge to $V(\pi) = 0$ (over $\pi \in \Pi$). □

## Appendix B: Proof of Theorem 3

Due to space constraints, this proof will be slightly dense. The proof techniques we use are due to Haussler [6] and Pollard [10]. Haussler [6], to which we will be repeatedly referring, provides a readable introduction to most of the methods used here.

We begin with some standard definitions from [6]. For a subset $T$ of a space $X$ endowed with (pseudo-)metric $\rho$, we say $T_0 \subset X$ is an $\epsilon$-**cover** for $T$ if, for every $t \in T$, there is some $t' \in T_0$ such that $\rho(t, t') \leq \epsilon$. For each $\epsilon > 0$, let $\mathcal{N}(\epsilon, T, \rho)$ denote the size of the smallest $\epsilon$-cover for $T$.

Let $\mathcal{H}$ be a family of functions mapping from a set $X$ into a bounded pseudo metric space $(A, \rho)$, and let $P$ be a probability measure on $X$. Define a pseudo metric on $\mathcal{H}$ by $d_{L^1(P,\rho)}(f, g) = \mathbf{E}_{x \sim P}[\rho(f(x), g(x))]$. Define the **capacity** of $\mathcal{H}$ to be $\mathcal{C}(\epsilon, \mathcal{H}, \rho) = \sup \mathcal{N}(\epsilon, \mathcal{H}, d_{L^1(P,\rho)})$, where the sup is over all probability measures $P$ on $X$. The quantity $\mathcal{C}(\epsilon, \mathcal{H}, \rho)$ thus measures the "richness" of the class $\mathcal{H}$. Note that $\mathcal{C}$ and $\mathcal{N}$ are both decreasing functions of $\epsilon$, and that $\mathcal{C}(\epsilon, \mathcal{H}, \rho) = \mathcal{C}(k\epsilon, \mathcal{H}, k\rho)$ for any $k > 0$.

The main results obtained with pseudo-dimension are uniform convergence of the empirical means of classes of random variables to their true means. Let $\mathcal{H}$ be a family of functions mapping from $X$ into $[0, M]$, and let $\vec{x}$ (the "training set") be $m$ i.i.d. draws from some probability measure $P$ over $X$. Then for each $h \in \mathcal{H}$, let $\hat{r}_h(\vec{x}) = (1/m) \sum_{i=1}^m h(x_i)$ be the empirical mean of $h(x)$. Also let $r_h(P) = \mathbf{E}_{x \sim P}[h(x)]$ be the true mean.

We now state a few results from [6]. In [6], these are Theorem 6 combined with Theorem 12; Lemma 7; Lemma 8; and Theorem 9 (with $Y$ being a singleton set, $\ell(y, a) = a$, $\alpha = \epsilon/4M$, and $\nu = 2M$). Below, $\ell_1$ and $\ell_2$ respectively denote the Manhattan and Euclidean metrics on $\mathbb{R}^n$. e.g. $\ell_1(\vec{x}, \vec{y}) = \sum_{i=1}^n |x_i - y_i|$.[11]

**Lemma 5** *Let $\mathcal{H}$ be a family of functions mapping from $X$ into $[0, M]$, and $d = \dim_P(\mathcal{H})$. Then for any probability measure $P$ on $X$ and any $0 < \epsilon \leq M$, we have that $\mathcal{N}(\epsilon, \mathcal{H}, d_{L^1(P,\ell_2)}) \leq 2((2eM/\epsilon) \ln(2eM/\epsilon))^d$.*

**Lemma 6** *Let $\mathcal{H}_1, \ldots, \mathcal{H}_k$ each be a family of functions mapping from $X$ into $[0, 1]$. The **free product** of the $\mathcal{H}_i$'s*

---

[11] This is inconsistent with the definition used in [6], which has an additional $(1/n)$ factor.

*is the class of functions $\mathcal{H} = \{(f_1, \ldots, f_k) : f_j \in \mathcal{H}_j\}$ mapping from $X$ into $[0, 1]^k$ (where $(f_1, \ldots, f_k)(x) = (f_1(x), \ldots, f_k(x))$). Then for any probability measure $P$ on $X$ and $\epsilon > 0$,*

$$\mathcal{N}(\epsilon, \mathcal{H}, d_{L^1(P,\ell_1)}) \leq \prod_{j=1}^k \mathcal{N}(\epsilon/k, \mathcal{H}_j, d_{L^1(P,\ell_2)}) \quad (10)$$

**Lemma 7** *Let $(X_1, \rho_1), \ldots, (X_{k+1}, \rho_{k+1})$ be bounded metric spaces, and for each $j = 1, \ldots, k$, let $\mathcal{H}_j$ be a class of functions mapping from $X_j$ into $X_{j+1}$. Suppose that each $\mathcal{H}_j$ is uniformly Lipschitz continuous (with respect to the metric $\rho_j$ on its domain, and $\rho_{j+1}$ on its range), with some Lipschitz bound $b_j \geq 1$. Let $\mathcal{H} = \{f_k \circ \cdots \circ f_1 : f_j \in \mathcal{H}_j, 1 \leq j \leq k\}$ be the class of functions mapping from $X_1$ into $X_{k+1}$ given by composition of the functions in the $\mathcal{H}_j$'s. Let $\epsilon_0 > 0$ be given, and let $\epsilon = k(\prod_{j=1}^k b_j)\epsilon_0$. Then*

$$\mathcal{C}(\epsilon, \mathcal{H}, \rho_{k+1}) \leq \prod_{j=1}^k \mathcal{C}(\epsilon_0, \mathcal{H}_j, \rho_{j+1}) \quad (11)$$

**Lemma 8** *Let $\mathcal{H}$ be a family of functions mapping from $X$ into $[0, M]$, and let $P$ be a probability measure on $X$. Let $\vec{x}$ be generated by $m$ independent draws from $X$, and assume $\epsilon > 0$. Then*

$$\Pr[\exists h \in \mathcal{H} : |\hat{r}_h(\vec{x}) - r_h(P)| > \epsilon]$$
$$\leq 4\mathcal{C}(\epsilon/16, \mathcal{H}, \ell_2) e^{-\epsilon^2 m / 64 M^2} \quad (12)$$

We are now ready to prove Theorem 3. No serious attempt has been made to tighten polynomial factors in the bound.

**Proof (of Theorem 3).** Our proof is in three parts. First, $\hat{V}$ gives an estimate of the discounted rewards summed over $(H_\epsilon + 1)$-steps; we reduce the problem of showing uniform convergence of $\hat{V}$ to one of proving that our estimates of the expected rewards on the $H$-th step, $H = 0, \ldots, H_\epsilon$, all converge uniformly. Second, we carefully define the mapping from the scenarios $s^{(i)}$ to the $H$-th step rewards, and use Lemmas 5, 6 and 7 to bound its capacity. Lastly, applying Lemma 8 gives our result. To simplify the notation in this proof, assume $R_{\max} = 1$, and $B, B_R \geq 1$.

**Part I: Reduction to uniform convergence of $H$-th step rewards.** $\hat{V}$ was defined by

$$\hat{V}(\pi) = \frac{1}{m} \sum_{i=1}^m R(s_0^{(i)}) + \gamma R(s_1^{(i)}) + \cdots + \gamma^{H_\epsilon} R(s_{H_\epsilon}^{(i)}).$$

For each $H$, let $\hat{V}_H(\pi) = \frac{1}{m} \sum_{i=1}^m R(s_H^{(i)})$ be the empirical mean of the reward on the $H$-th step, and let $V_H(\pi) = \mathbf{E}_{s_H}[R(s_H)]$ be the true expected reward on the $H$-th step (starting from $s_0 \sim D$ and executing $\pi$). Thus, $V(\pi) = \sum_{H=0}^\infty \gamma^H V_H(\pi)$.

Suppose we can show, for each $H = 0, \ldots, H_\epsilon$, that with probability $1 - \delta/(H_\epsilon + 1)$,

$$|\hat{V}_H(\pi) - V_H(\pi)| \leq \epsilon/2(H_\epsilon + 1) \quad \forall \pi \in \Pi \quad (13)$$

Then by the union bound, we know that with probability $1 - \delta$, $|\hat{V}_H(\pi) - V_H(\pi)| \leq \epsilon/2(H_\epsilon + 1)$ holds simultaneously for all $H = 0, \ldots, H_\epsilon$ and for all $\pi \in \Pi$. This implies that, for all $\pi \in \Pi$,

$$|\hat{V}(\pi) - V(\pi)|$$
$$\leq |\hat{V}(\pi) - \sum_{H=0}^{H_\epsilon} \gamma^H V_H(\pi)| + |\sum_{H=0}^{H_\epsilon} \gamma^H V_H(\pi) - V(\pi)|$$
$$\leq \sum_{H=0}^{H_\epsilon} |\hat{V}_H(\pi) - V_H(\pi)| + \epsilon/2$$
$$\leq \epsilon.$$

where we used the fact that $|\sum_{H=0}^{H_\epsilon} \gamma^H V_H(\pi) - V(\pi)| \leq \epsilon/2$, by construction of the $\epsilon$-horizon time. But this is exactly the desired result. Thus, we need only prove that Equation (13) holds with high probability for each $H = 0, \ldots, H_\epsilon$.

**Part II: Bounding the capacity.** Let $H \leq H_\epsilon$ be fixed. We now write out the mapping from a scenario $s^{(i)} \in S \times ([0,1]^{d_P})^\infty$ to the $H$-th step reward. Since this mapping depends only on the first $d_P \cdot H$ elements of the "$p$"s portion of the scenario, we will, with some abuse of notation, write the scenario as $s^{(i)} \in S \times [0,1]^{d_P H}$, and ignore its other coordinates. Thus, a scenario $s^{(i)}$ may now be written as $(s, p_1, p_2, \ldots, p_{d_P H})$.

Given a family of functions (such as $\mathcal{F}_i$) mapping from $S \times [0,1]^{d_P}$ into $[0,1]$, we extend its domain to $S \times [0,1]^{d_P + n}$ for any finite $n \geq 0$ simply by having it ignore the extra coordinates. Note this extension of the domain does not change the pseudo-dimension of a family of functions. Also, for each $n = 1, \ldots, n$, define a mapping $I_n$ from $S \times [0,1]^n \mapsto [0,1]$ according to $I_n(s, p_1, p_2, \ldots, p_n) = p_n$. For each $n$, let $\mathcal{I}_n = \{I_n\}$ be singleton sets. Where necessary, $I_n$'s domain is also extended as we have just described.

For each $i = 1, \ldots, H+1$, define $X_i = S \times ([0,1]^{d_P})^{H+1-i}$. For example, $X_1$ is just the space of scenarios (with only the first $d_P H$ elements of the $p$'s kept), and $X_{H+1} = S$. For each $i = 1, \ldots, H$, define a family of maps from $X_i$ into $X_{i+1}$ according to $\mathcal{H}_i = \mathcal{F}_1 \times \mathcal{F}_2 \times \cdots \times \mathcal{F}_{d_S} \times \mathcal{I}_{d_P+1} \times \mathcal{I}_{d_P+2} \times \cdots \times \mathcal{I}_{(H-i+1)d_P}$ (where the definition of the free product of sets of functions is as given in Lemma 6); note such an $\mathcal{H}_i$ has Lipschitz bound at most $B_0 = (d_S + H d_P)B$. Also let $\mathcal{H}_{H+1} = \{R\}$ be a singleton set containing the reward function, and $X_{H+2} = [-R_{\max}, R_{\max}]$. Finally, let $\mathcal{H} = \mathcal{H}_{H+1} \circ \mathcal{H}_H \circ \cdots \circ \mathcal{H}_1$ be the family of maps from $S \times ([0,1]^{d_P})^H$ into $[-R_{\max}, R_{\max}]$.

Now, let $\hat{V}^\pi_{M',H} : S' \mapsto [-R_{\max}, R_{\max}]$ be the reward received on the $H$-th step when executing $\pi$ from a scenario $s \in S'$. As we let $\pi$ vary over $\Pi$, this defines a family of maps from scenarios into $[-R_{\max}, R_{\max}]$. Clearly, this family of maps is a subset of $\mathcal{H}$. Thus, if we can bound the capacity of $\mathcal{H}$ (and hence prove uniform converge over $\mathcal{H}$), we have also proved uniform convergence for $\hat{V}^\pi_{M',H}$ (over all $\pi \in \Pi$).

For each $i = 1, \ldots, d_S$, since $\dim_P(F_i) \leq d$, Lemma 5 implies that $\mathcal{N}(\epsilon, \mathcal{F}_i, d_{L^1(P,\ell_2)}) \leq 2((2e/\epsilon)\ln(2e/\epsilon))^d$. Moreover, clearly $\mathcal{N}(\epsilon, \mathcal{I}_i, d_{L^1(P,\ell_2)}) = 1$ since each $\mathcal{I}_i$ is a singleton set. Combined with Lemma 6, this implies that, for each $i = 1, \ldots, H$ and $\epsilon \leq 1$,

$$\mathcal{N}(\epsilon, \mathcal{H}_i, d_{L^1(P,\ell_1)})$$
$$\leq \prod_{j=1}^{d_S} \mathcal{N}(\epsilon/(d_S + (H-i)d_P), \mathcal{F}_j, d_{L^1(P,\ell_2)})$$
$$\leq \prod_{j=1}^{d_S} \mathcal{N}(\epsilon/(d_S + H_\epsilon d_P), \mathcal{F}_j, d_{L^1(P,\ell_2)})$$
$$\leq 2^{d_S} \left(\frac{2e(d_S + H_\epsilon d_P)}{\epsilon} \ln \frac{2e(d_S + H_\epsilon d_P)}{\epsilon}\right)^{dd_S}$$
$$\leq 2^{d_S} \left(\frac{2e(d_S + H_\epsilon d_P)}{\epsilon}\right)^{2dd_S}$$

where we have used the fact that $\mathcal{N}$ is decreasing in its $\epsilon$ parameter. By taking a sup over probability measures $P$, this is also a bound on $\mathcal{C}(\epsilon, \mathcal{H}_i, \ell_1)$. Now, as metrics over $\mathbb{R}^{(d_S + (H-i)d_P)}$, $\ell_2 \leq \ell_1$. Thus, this also gives

$$\mathcal{C}(\epsilon, \mathcal{H}_i, \ell_2) \leq 2^{d_S} \left(\frac{2e(d_S + H_\epsilon d_P)}{\epsilon}\right)^{2dd_S} \quad (14)$$

Finally, applying Lemma 7 with each of the $\rho_k$'s being the $\ell_2$ norm on the appropriate space, $k = H + 1$, and $\epsilon = (H+1)B_0^H B_R \epsilon_0$, we find

$$\mathcal{C}(\epsilon, \mathcal{H}, \ell_2)$$
$$\leq \prod_{j=1}^{H+1} \mathcal{C}(\epsilon/((H+1)B_0^H B_R), \mathcal{H}_j, \ell_2)$$
$$\leq \prod_{j=1}^{H} 2^{d_S} \left(\frac{2e(d_S + H_\epsilon d_P)(H+1)B_0^H B_R}{\epsilon}\right)^{2dd_S}$$
$$\leq 2^{d_S H_\epsilon} \left(\frac{2e(d_S + H_\epsilon d_P)(H_\epsilon + 1)B_0^{H_\epsilon} B_R}{\epsilon}\right)^{2dd_S H_\epsilon}$$

**Part III: Proving uniform convergence.** Applying Lemma 8 with the above bound on $\mathcal{C}(\epsilon, \mathcal{H}, \ell_2)$, we find that for there to be a $1 - \delta$ probability of our estimate of the expected $H$-th step reward to be $\epsilon$-close to the mean, it suffices that

$$m = \frac{256}{\epsilon^2}\left(\log \frac{1}{\delta} + \log(4\mathcal{C}(\epsilon/16, \mathcal{H}, \ell_2))\right)$$
$$= O\left(\text{poly}\left(d, \frac{1}{\epsilon}, \log \frac{1}{\delta}, \frac{1}{1-\gamma}, \log B, \log B_R, d_S, d_P\right)\right).$$

This completes the proof of the Theorem. □